\documentclass{article}
\usepackage{spconf,amsmath,graphicx,epsfig,multirow}

\title{The 2015 Sheffield System for Transcription of Multi--Genre \\ Broadcast Media}
\name{{\em Oscar Saz, Mortaza Doulaty, Salil Deena, Rosanna Milner,}\\{\em Raymond W.M. Ng, Madina Hasan, Yulan Liu, Thomas Hain}\thanks{This work was supported by the EPSRC Programme Grant EP/I031022/1 (Natural Speech Technology).}}
\address{Speech and Hearing Group, Department of Computer Science, University of Sheffield, UK}
\begin{document}
\ninept
\maketitle
\begin{abstract}
We describe the University of Sheffield system for participation in the 2015 Multi--Genre Broadcast 
(MGB) challenge task of transcribing multi--genre broadcast shows. Transcription was one of four 
tasks proposed in the MGB challenge, with the aim of advancing the state of the art of automatic 
speech recognition, speaker diarisation and automatic alignment of subtitles for  broadcast 
media. Four topics are investigated in this work: Data selection techniques for training with 
unreliable data, automatic speech segmentation of broadcast media shows, acoustic modelling and 
adaptation in highly variable environments, and language modelling of multi--genre shows. The final 
system operates in multiple passes, using an initial unadapted decoding stage to refine 
segmentation, followed by three adapted passes: a hybrid DNN pass with input features normalised by 
speaker--based cepstral normalisation, another hybrid stage with input features normalised by 
speaker feature--MLLR transformations, and finally a bottleneck--based tandem stage with noise and 
speaker factorisation. The combination of these three system outputs provides a final error rate of 
27.5\% on the official development set, consisting of  47 multi--genre shows. 
\end{abstract}
\section{Introduction}
\label{sec:intro}


Audio-visual media is an area of high interest for research in a variety of topics related to 
computer vision, speech processing and natural language processing. The ability to search into vast 
media archives, browse through thousands of hours of recordings or structure the complete resources 
of a media company would significantly increase the efficiency of these organisations and the 
services provided to the end users.

From the point of view of Automatic Speech Recognition (ASR), work on transcription of broadcast 
news has achieved significant reduction in error rates since the early works in the 1990s 
\cite{Woodland97, Gauvain02}, with word error rates falling below 10\% for traditional broadcast new 
programmes \cite{Gales06}. However, other types of broadcast media shows have not been so 
widely explored. The transcription of multi-genre data is a complex task due to the large amounts of 
variability arising from multiple, diverse speakers, the variety of acoustic and recording 
conditions and the lexical and linguistic diversity of the topics covered \cite{Lanchantin13}.

Evaluations of technology covering different aspects of research in audio-visual media have been a 
major driver behind some of the most recently achieved results in audio-visual media processing. 
The MediaEval evaluation campaign \cite{MediaEval} has brought together researchers from many 
areas to work in automatic classification and retrieval of broadcast data. Evaluation series such 
as the NIST-organised Hub4 tasks \cite{Hub4} helped start the earlier efforts in broadcast news 
transcriptions in English, while the Topic Detection and Tracking (TDT) campaign \cite{TDT} 
expanded this work to other tasks related to broadcast news. More recently, the Ester campaigns 
\cite{Ester} have created increased interest in the transcription of French broadcast news and the 
Albayzin campaigns \cite{Albayzin} have pushed the efforts in audio processing of Spanish broadcast 
news.

Following these efforts, the Multi-Genre Broadcast (MGB) challenge \cite{MGB} aimed to take on 
several tasks of an increasing complexity in broadcast media. This work tries to address that with 
advances in several areas of ASR and its application in a fully functional system for Task 1 of the 
MGB challenge: Speech-to-text transcription of broadcast television.

The rest of the paper is organised as follows: Section \ref{sec:setup} describes the 
experimental setup. Section \ref{sec:train} explains data selection techniques used for acoustic 
model training. Section \ref{sec:seg} introduces new procedures for improved automatic segmentation 
for ASR. Sections \ref{sec:varia} and \ref{sec:lm} describe different approaches for acoustic 
model adaptation and language modelling adaptation for multi-genre shows. Section \ref{sec:system} 
outlines the final system. Overall results are presented in Section \ref{sec:results}. Finally, 
Section \ref{sec:conclusion} discusses outcomes and concludes the paper.

\section{MGB Challenge - Task 1}
\label{sec:setup}

The MGB challenge 2015 consisted of four different tasks, covering the topics of multi-genre 
broadcast show transcription, lightly supervised alignment, longitudinal broadcast transcription 
and longitudinal speaker diarisation. The focus of this work was on Task 1: Speech-to-text 
transcription of broadcast television, although aspects of the system presented here were used 
in submissions to other challenge tasks. A full description of this and the other tasks in the 
challenge can be found in \cite{MGB}, but a brief description of the task is given here.

Participation in this task required the automatic transcription of a set of shows 
broadcast by the British Broadcasting Corporation (BBC). These shows were chosen to cover the 
multiple genres in broadcast TV, categorised in terms of 8 genres: advice, children's, comedy, 
competition, documentary, drama, events and news. Acoustic Model (AM) training data was fixed and 
limited to more than 2,000 shows, broadcast by the BBC during 6 weeks in April and May of 2008. The 
development data for the task consisted of 47 shows that were broadcast by the BBC during a week in 
mid-May 2008. The numbers of shows and the associated broadcast time for training and development 
data are shown in Table \ref{tab:mgbdata}.

\begin{table} [th]
\caption{\label{tab:mgbdata} {\it Amount of training and development data.}}
\centerline{
\begin{tabular}{|c|c|c|c|c|}
\hline
 &  \multicolumn{2}{|c|}{Train} & \multicolumn{2}{|c|}{Development} \\
\hline 
Genre & Shows & Time & Shows & Time \\
\hline
Advice & 264 & 193.1h. & 4 & 3.0h. \\
Children's & 415 & 168.6h. & 8 & 3.0h. \\
Comedy & 148 & 74.0h. & 6 & 3.2h. \\
Competition & 270 & 186.3h. & 6 & 3.3h. \\
Documentary & 285 & 214.2h. & 9 & 6.8h. \\
Drama & 145 & 107.9h. & 4 & 2.7h. \\
Events & 179 & 282.0h. & 5 & 4.3h. \\
News & 487 & 354.4h. & 5 & 2.0h. \\
\hline
Total & 2,193 & 1580.5h.  & 47 & 28.3h.  \\
\hline
\end{tabular}}
\end{table}
\vspace{-3mm}

Additional data was available for Language Model (LM) training in the form of subtitles from shows 
broadcast since 1979 to March 2008, with a total of 650 million words, and referred to as $LM1$. The 
subtitles from the 2,000+ shows for acoustic modelling could also be used for LM training, referred 
to as $LM2$. Statistics for these 2 sets can be seen in Table \ref{tab:mgbdata2}. 


\vspace{-5mm}
\begin{table} [th]
\caption{\label{tab:mgbdata2} {\it Amount of language model training data.}}
\centerline{
\begin{tabular}{|c|c|c|c|}
\hline
Subtitles &  \#sentences & \#words & \#unique words \\
\hline 
$LM1$ (1979-2008) & 72.9M & 648.0M & 752,875 \\
$LM2$ (Apr/May '08) & 633,634 & 10.6M & 32,304 \\
\hline
\end{tabular}}
\end{table}
\vspace{-5mm}

\subsection{Common system description}
\label{ssec:base}

Throughout this work, two different types of systems were used. This Section describes the 
fundamental features for both of them, while specific descriptions will be given in the paper, if
further experiments are addressing specific issues.


The first types of systems used were $Hybrid$ DNN-HMM systems, built using the Kaldi toolkit 
\cite{Kaldi}. These were based on a Deep Neural Network (DNN) where the input were 5 contiguous 
spliced frames of Perceptual Linear Prediction (PLP) features of $40$ dimensions. Features
were obtained by using a linear discriminant analysis transformation of 117 spliced PLP features 
(from 13 dimensions with a context of 4 to the left and right and middle frame), followed by a global 
CMLLR transform. Features were transformed using a boosted Maximum Mutual Information (bMMI) 
discriminative transformation \cite{PoveyKKRSV08}, unless otherwise stated. DNNs consisted of 6 
hidden layers of 2,048 neurons, and an output layer of 6,478 triphone state targets. State-level 
Minimum Bayes Risk (sMBR) \cite{KingsburySS12,gibson_is06} as target functions, unless otherwise 
mentioned, and Stochastic Gradient Descent (SGD) was used as the optimisation method. Decoding with 
$Hybrid$ systems was performed in two stages; in the first stage, lattices were generated using a 
highly pruned 3-gram, and afterwards the lattices were rescored using a complete 4-gram and the 
1-best obtained and scored using the official MGB scoring package.

The second system types used are so-called $Bottleneck$ DNN-GMM-HMM systems built using the TNet 
toolkit \cite{TNet} for DNN training and the HTK toolkit \cite{HTK} for Gaussian Mixture Model (GMM) 
and Hidden Markov Model (HMM) training and decoding. $Bottleneck$ systems used a DNN as a front-end 
for extracting a set of 26 bottleneck features. Such DNNs took as input 15 contiguous log-filterbank 
frames and consisted of 4 hidden layers of 1,745 neurons plus the 26-neuron bottleneck layer, and an 
output layer of 8,000 triphone state targets. sMBR was used for training, unless otherwise stated. 
Feature vectors for training the GMM-HMM systems were 65-dimensional, including the 26 dimensional 
bottleneck features, as well as 13 dimensional PLP features together with their first and second 
derivatives. GMM-HMM models were trained using 16 Gaussian components per state, and around 8k 
distinct triphone states. Decoding with $Bottleneck$ systems was also performed in two stages; in a 
first stage, lattices were generated using a 2-gram, and afterwards these lattices were rescored 
using a 4-gram and the 1-best obtained and scored with the official MGB scoring package.

All decoding experiments were performed using a 50,000-word vocabulary, constructed from the most 
frequent words in the subtitles as provided for language model training. Pronunciations were 
obtained using the Combilex pronunciation dictionary\cite{Combilex}, which was provided to the 
challenge participants. When a certain word was not contained in the lexicon, automatically 
generated pronunciations were obtained using the Phonetisaurus toolkit \cite{Phonetisaurus}. These 
pronunciations were expanded to incorporate pronunciation probabilities, learnt from the alignment 
of the AM training data \cite{Hain05}. Unless otherwise stated, language models used were 
obtained by interpolation of several language models trained with the $LM1$ and 
$LM2$ language model data from Table \ref{tab:mgbdata2}. LM training was performed with the SRILM 
toolkit \cite{SRILM}.

\section{Data Selection and Training}
\label{sec:train}

One of the main difficulties for transcription in the MGB challenge was the efficient use of the 
acoustic training data provided, as the use of prior models or other data was not allowed. The 
transcription of the training data was not created for ASR training purposes.  Only the subtitle 
text broadcast with each show could be used, which is of varying quality for a variety of reasons. 
An aligned version of the subtitles was provided where the time stamps of the subtitles had been 
corrected in a lightly supervised manner \cite{MGB,Long13}. After this process, 1,196.73 hours of 
speech were left available for training.

The provided transcripts for the training shows were unreliable in two ways: First, the 
subtitle text might not always match the actual spoken words; and second, the time boundaries given 
might have errors arising from the lightly supervised alignment process. This work did not aim to 
improve on the second aspect, but instead it studied how to perform data selection in order to train 
with those segments with the most accurate transcripts. An initial selection strategy was based on 
selecting segments for training based on their Word Matching Error Rate (WMER), a by-product of the 
semi-supervised alignment process that measures how similar the text in the subtitle matched the 
output of a lightly supervised ASR system for that segment \cite{MGB,Long13}.

A more complex selection strategy was designed using confidence scores for each segment. The scores 
were obtained from the posterior probabilities given by a 4-layer DNN trained on the initial 
selection of data whose targets were 144 monophone states \cite{Zhang14}. The inputs to this DNN were 
15 contiguous log-Mel-filter-bank frames, and each hidden layer had 1,745 neurons. For each segment 
in the training set, the monophone state sequence was obtained using forced alignment, and the 
segment-based confidence measure was calculated as the average of the logarithmic posteriors of each 
frame for its corresponding monophone state, excluding silence areas.

Two different training data setups arose from these two strategies: $TRN1$, which contained 512.6 
hours of speech segments with WMER of 40\% or less; and $TRN2$, which contained 698.9 hours of 
speech segments with confidence score above $-3.0$. The amount of data per genre in each data 
training definition can be found in Table \ref{tab:datasel}.

\begin{table} [th]
\caption{\label{tab:datasel} {\it $TRN1$ and $TRN2$ data selection strategies.}}
\centerline{
\begin{tabular}{|c|c|c|}
\hline
 &  $TRN1$ & $TRN2$ \\
\hline
Advice & 72.2h. & 107.8h. \\
Children's & 54.2h. & 68.9h. \\
Comedy & 17.3h. & 26.2h. \\
Competition & 68.5h. & 99.0h. \\
Documentary & 92.6h. & 113.5h. \\
Drama & 24.1h. & 36.3h. \\
Events & 34.2h. & 44.1h. \\
News & 153.4h. & 203.0h. \\
\hline
Total & 512.5h.  & 698.8h.  \\
\hline
\end{tabular}}
\end{table}

Both training strategies were evaluated on the $Hybrid$ and $Bottleneck$ systems, as defined in 
Section \ref{ssec:base}, in this case using Cross-Entropy (CE) training \cite{Hinton12}. 
Recognition experiments were performed on the manual segmentation available for the development 
data, with the Word Error Rate (WER) results shown in Table \ref{tab:datasel2}. The results indicate
that there is a 1\% absolute improvement from using $TRN2$ instead of $TRN1$, although the gain 
might have been due mainly to the extra 180 hours of data included in $TRN2$. The gain was 
independent of the system setup, and was achieved in both $Hybrid$ and $Bottleneck$ 
systems.

\begin{table} [th]
\caption{\label{tab:datasel2} {\it ASR results with different data selection strategies.}}
\centerline{
\begin{tabular}{|c|c|c|c|}
\hline
System & Training data & WER\\
\hline
\multirow{2}{*}{$Hybrid$} & $TRN1$ & 30.6\% \\
 & $TRN2$  & 29.0\% \\
 \hline
 \multirow{2}{*}{$Bottleneck$} & $TRN1$ & 34.4\% \\
  & $TRN2$  & 33.3\% \\
\hline
\end{tabular}}
\end{table}

\section{Automatic Segmentation}
\label{sec:seg}

Automatic speech segmentation is a very important aspect in automatic processing of broadcast media, 
where the presence of music, applause, laughter and other background sounds can significantly 
degrade the ability to detect sections containing speech. Errors in segmentation can then propagate 
as ASR errors in regions of undetected speech or those where speech was incorrectly detected. In 
this work, a multi-stage automatic segmentation procedure is introduced: an initial segmentation 
based on DNN posteriors is subsequently improved using the output of an ASR system.

NNs have been used extensively for speech segmentation of meetings \cite{dines_is06,Hain2012} and 
naturally DNNs are equally useful for this task \cite{Ryant13}. The neural networks are trained to 
classify each frame in one of two classes, one corresponding to speech being present and the other 
one representing speech not being present. One of the challenges in this work's setup was, as seen 
in the previous section, the unreliability of the data and the requirement to have efficient data 
selection strategies. Two strategies were tested to cope with the issue. In the first one, $SNS1$, 
all acoustic training data available were used for training the DNN, the originally defined segments 
were force-aligned to determine which areas were speech and which areas were non-speech. All audio 
that was not assigned to a speech segment in the original segments was labelled as non-speech. The 
second strategy, $SNS2$, took the 512.5 hours from the $TRN1$ data selection strategy, as defined 
in Section \ref{sec:train}, and used force alignment to label areas as speech and non-speech, without 
adding any extra non-speech areas. The amount of training data can be seen in Table \ref{tab:vad2}.

\begin{table} [th]
\caption{\label{tab:vad2} {\it $SNS1$ and $SNS2$ data selection strategies for speech segmentation.}}
\centerline{
\begin{tabular}{|c|c|c|c|}
\hline
 &  Speech & Non-speech & Total \\
\hline
$SNS1$ & 759h. & 793h. & 1,552h. \\
$SNS2$ & 363h. & 116h. & 479h. \\
\hline
\end{tabular}}
\end{table}

\begin{table*} [th]
\caption{\label{tab:vad} {\it Results in automatic segmentation.}}
\centerline{
\begin{tabular}{|c|c|c|c|c|c|c|}
\hline
 & Speech time & Segments & Missed speech & False speech & $Hybrid$ WER  & $Bottleneck$ WER \\
\hline
Human & 19.5h. & 30,702 & 0.0\% & 0.0\%  & 28.6\% & 31.0\%\\
$SNS1$ & 18.3h. & 17,713 & 6.6\% & 2.6\% & 31.2\% & 34.4\%\\
$SNS2$ & 21.8h. & 15,337 & 1.3\% & 15.4\% & 31.0\% & 34.4\%\\
 +Refinement & 19.3h. & 16,327 &  4.0\% & 5.4\% & 29.8\% & 33.3\% \\
\hline
\end{tabular}}
\end{table*}

The segmentation DNN provided, for any given audio output, the estimated values of the posterior 
probabilities of speech or non-speech for each frame. A two-state HMM was used to smooth this 
sequence of posteriors to a sequence of valid speech segments, with extra 0.25 seconds added at the 
beginning and the end of each speech segment. This, with either of the strategies $SNS1$ or $SNS2$, 
gave the initial segmentation used for recognition in the first pass.

With the output of decoding based on the original segmentations, a refinement stage was performed as 
follows. Confidence measures based on the posteriors of a 144-monophone-target DNN were obtained for 
each word in the hypothesis, as seen for acoustic data selection in section \ref{sec:train}. Then, 
the raw confidence scores were mapped using a decision tree trained on the development data, using 
decision targets that were either $1$ if the word was in an area of speech as defined in the 
reference segmentation, or $0$ if the word was in an area of non-speech. The features to the 
decision tree were the raw confidence score of each word, the confidence score of the segment, the 
length of the word (in seconds), the length of the word (in phonemes) and the length of the segment 
(in seconds). Once the confidences were calculated, words with confidence score below a threshold 
were removed from the transcript. New segments were redefined then around the remaining words.

The results of the this systems are presented in Table \ref{tab:vad}, in terms of segmentation 
error: i.e. missed speech and false alarms, and WER for sMBR $Hybrid$ and $Bottleneck$ systems 
trained on the $TRN2$ data. Both DNN segmenters produced a significant degradation compared to the 
use of manually defined segments. However, $SNS2$ was found to achieve a much larger false alarm 
rate than $SNS1$, possibly due to the unbalanced amount of data used for training $SNS2$. This made 
$SNS2$ more suitable for the refinement stage, where areas of false speech detection could be pruned 
by the use of confidence measures in the ASR output. Table \ref{tab:vad} shows how this refinement 
stage using ASR gave more than 1\% absolute improvement over $SNS1$ and $SNS2$, despite its 
segmentation error rate of 9.4\%, similar to $SNS1$ at 9.2\%.

\section{Acoustic Background Modelling}
\label{sec:varia}

Tackling acoustic variability is one of the main issues arising for multi-genre broadcast 
transcription. The presence of a large variety of possible recording conditions and acoustic 
background environments presents a real challenge for ASR systems. In this work, two approaches to 
compensating for such variability were studied. The first aimed to normalise the background 
variability in the input to DNNs for hybrid systems, while the second one aimed to use asynchronous 
Constrained Maximum Likelihood Linear Regression (aCMLLR) transformations \cite{Saz13} for the 
compensation of dynamic background noises in bottleneck systems.

\begin{table*} [th]
\caption{\label{tab:adapt} {\it Domain and noise adaptation of $Hybrid$ and $Bottleneck$ systems}}
\centerline{
\begin{tabular}{|c|c|c|c|c|c|c|c|c|c|}
\hline
System & Adv. & Child. & Comed. & Compet. & Docum. & Dram. & Even. & News & Global \\
\hline
$Hybrid$ CE baseline              & 26.9\% & 26.8\% & 45.9\% & 25.5\% & 28.5\% & 49.1\% & 33.0\% & 16.1\% & 30.7\% \\
$Hybrid$ CE adapted                & 24.2\% & 26.5\% & 43.8\% & 23.6\% & 27.3\% & 45.0\% & 31.6\% & 14.3\% & 28.9\% \\
\hline
$Bottleneck$ baseline           & 25.2\% & 30.8\% & 44.7\% & 27.3\% & 28.9\% & 42.1\% & 34.9\% & 16.6\% & 31.0\% \\
$Bottleneck$ adapted            & 24.6\% & 29.2\% & 43.3\% & 26.7\% & 27.9\% & 40.8\% & 33.8\% & 15.8\% & 30.0\% \\
\hline
\end{tabular}}
\end{table*}

\subsection{Domain adaptation of hybrid systems}

Adaptation of DNN-based ASR systems is currently one of the most extensively researched areas of 
speech recognition technology. While several approaches have been evaluated in the past, the 
normalisation of the input features is most commonly employed. For example, for speaker adaptation, 
this has been done by directly transforming the input features via feature MLLR (fMLLR) 
transformations \cite{Gales1998} or by using additional input features representing some 
characteristic of the speaker, like i-Vectors \cite{Karanasou15,Liu15}.

Latent Dirichlet Allocation (LDA) models have been recently used to model hidden acoustic categories 
in audio data. In \cite{Doulaty15}, it was shown that LDA is a suitable model for structuring 
acoustic data from unknown origin, into unsupervised categories, that could be used to provide 
domain adaptation in ASR. In this work, 64 hidden acoustic domains were found in the acoustic model
training data using the LDA model following the procedure in \cite{Doulaty15}; these domains were found in a unsupervised
manner and internally structured the different acoustic conditions of the data. Afterwards, each segment in the
training and development sets was assigned to one of these domains. In DNN training, 64 extra features
were appended in the input layer, where the domain corresponding to the input frame was codified as a 1--of--N vector.
Decoding is performed as usual, with the hidden domain corresponding to the input segment being also appended in
the input layer.

\subsection{Dynamic noise adaptation of bottleneck systems}

One of the advantages of tandem (DNN-GMM-HMM) systems is that techniques for adaptation such as 
Maximum A Posteriori (MAP) or MLLR \cite{Gales96} can be employed. In our previous works, a new HMM 
topology for asynchronous adaptation of GMM-HMM systems was proposed and shown to produce ASR 
improvement in the presence of dynamic background conditions \cite{Saz13}.

This setup was applied to this task and expanded through the use of asynchronous Noise Adaptive 
Training (aNAT) \cite{Kalinli10, Saz13}. First, a global aCMLLR transformation with 8 parallel paths 
was trained on the whole training data in order to characterise the most common background 
conditions in this data. Then, the initial sMBR-trained $Bottleneck$ model was retrained in an 
adaptive training fashion using this aCMLLR transformation. Finally, the global aCMLLR 
transformation was retrained into show-based aCMLLR transformations using an initial decoding stage 
in order to more finely characterise the types of noise and background existing in each show, and 
these transformations were used with the aNAT $Bottleneck$ model to run the final noise-adapted 
system.

The results, including baseline results, for $Hybrid$ systems with domain adaptation and 
$Bottleneck$ systems with noise adaptation are shown in Table \ref{tab:adapt} using the manually 
defined segmentation and for systems trained on $TRN2$ data. The $Hybrid$ baseline and $Hybrid$ 
adapted systems were cross-entropy (CE) trained in this case, because sequence training for 
domain-adapted hybrid DNNs did not complete in time. The domain adapted DNN in the $Hybrid$ setup provided a significant
improvement of 1.8\% (5.9\% relative), which showed the strength of the hidden domain found through the LDA model.
For $Bottleneck$ systems, the improvement over 
the baseline was 1\% absolute (3.2\% relative) in WER, with balanced improvement across the 8 
genres. The experiments in Table \ref{tab:adapt} were carried out after the challenge and thus were 
not a part of the final submission.

\section{Multi--genre Language Modelling}
\label{sec:lm}

\begin{table*} [th]
\caption{\label{tab:lm} {\it Results with genre-LMs for $Hybrid$ and $Bottleneck$ systems}}
\centerline{
\begin{tabular}{|c|c|c|c|c|c|c|c|c|c|c|}
\hline
\multicolumn{2}{|c|}{System}  & Adv. & Child. & Comed. & Compet. & Docum. & Dram. & Even. & News & Global \\
\hline
\multirow{2}{*}{$Hybrid$ baseline 4-gram LM} & PPL & 94.5 & 101.4 & 102.1 & 104.2 & 129.4 & 83.9 & 126.3 & 137.1 & 110.8\\
 &          WER & 23.6\% & 27.9\% & 41.1\% & 25.3\% & 27.2\% & 38.2\% & 33.4\% & 15.3\% & 28.9\% \\
 \multirow{2}{*}{$Hybrid$ genre RNN LM} & PPL & 58.6 & 62.7 & 59.6 & 50.5 & 68.7 & 60.4 & 64.0 & 67.2 & N/A\\
 & WER & 23.0\% & 23.6\% & 43.9\% & 22.7\% & 27.3\% & 43.9\% & 31.7\% & 13.6\% & 28.2\% \\
\hline
\multirow{2}{*}{$Bottleneck$ baseline 4-gram LM} & PPL & 94.5 & 101.4 & 102.1 & 104.2 & 129.4 & 83.9 & 126.3 & 137.1 & 110.8\\
 & WER & 25.2\% & 30.8\% & 44.7\% & 27.3\% & 28.9\% & 42.1\% & 34.9\% & 16.6\% & 31.0\% \\
 \multirow{2}{*}{$Bottleneck$ genre 4-gram LM} & PPL & 87.2 & 92.1 & 93.8 & 94.5 & 124.1 & 78.4 & 120.0 & 125.1 & N/A\\
 & WER & 24.9\% & 30.6\% & 44.3\% & 27.0\% & 28.7\% & 41.8\% & 34.9\% & 16.2\% & 30.8\% \\
\hline
\end{tabular}}
\end{table*}

Acoustic variation is not the only source of variability that can be found in multi-genre 
broadcasts. Lexical and linguistic variability is also present in this data, due to the large 
variety of topics that are covered in these shows. In order to tackle this linguistic variability, 
several experiments were designed to improve language modelling in this task.

One of the aspects explored in this work is the use of genre-specific LMs. While the subtitles 
in the $LM2$ language model training data were already categorised by genre, this information was 
not available in the much larger $LM1$ language model training data. In order to automatically 
derive genre labels for that dataset, genres were automatically inferred using an LDA based 
approach. First, hidden LDA topics were inferred from the $LM2$ data where genre labels are present. 
Given those, a Support Vector Machines (SVM) classifier could be trained that would allow classifying
a show into one of the 8 genres using the distribution of LDA hidden topic posteriors as 
input. These SVMs were used to produce labels for separated chunks of the $LM1$ training data. The 
statistics of words assigned to each genre can be seen in Table \ref{tab:lda}.

\begin{table} [th]
\caption{\label{tab:lda} {\it Number of words for training of genre-LMs.}}
\centerline{
\begin{tabular}{|c|c|c|}
\hline
 &  $LM1$ & $LM2$ \\
\hline
Advice & 91.8M & 1.4M \\
Children's & 41.4M & 0.8M \\
Comedy & 98.5M & 0.4M \\
Competition & 73.6M & 1.3M \\
Documentary & 189.2M & 1.2M \\
Drama & 97.7M & 0.5M \\
Events & 4.4M & 1.1M \\
News & 51.9M & 2.7M \\
\hline
\end{tabular}}
\end{table}


Once all the data had been classified into genres, genre-based LMs were trained in two different 
configurations: The first one was based on a Recurrent Neural Network (RNN) LM
\cite{Mikolov10}, initially trained on the full $LM1$ and $LM2$ training data. This initial RNNLM was then converted into 8 genre--dependent RNNLMs by fine--tuning each one of them to the genre-dependent data. These RNNLMs were 
used to rescore the lattices obtained by the $Hybrid$ systems using the baseline 4-gram language model. 
The second one was based on genre-based 4-grams as the interpolation of the genre-independent 4-gram 
with each genre-dependent 4-gram and was used to rescore lattices in $Bottleneck$ systems. Both 
systems used manual segmentation and were trained on $TRN2$.

The perplexity and recognition results obtained with  the genre-specific LMs are shown in Table 
\ref{tab:lm}, along with the results using the baseline LMs. The results show a very 
significant drop in perplexity when using RNNLMs but only a modest improvement in word error rate of 
0.7\%. This is consistent with the experiments reported on the same BBC data in \cite{ChenTLLWGW15}. 
The main difference, however is that in \cite{ChenTLLWGW15}, instead of $LM1$ as background language 
model, another corpus of 1 billion words was used for language modelling, and different topic 
models including LDA, were used to classify the text into a set of different genres. As noted 
above, the LM training data is noisy, both in word accuracy and genre labelling. 

Using genre-specific n-gram language models yields an improvement of only 0.2\% and the perplexity 
reductions are not as significant. This could be explained by the need to use longer contexts than 
4-grams, in order to obtain improvements, which RNNLMs are able to achieve through the use of 
unrestrained context. It is also interesting to note that genre-specific RNNLMs perform worse than 
corresponding n-grams on some genres (e.g., comedy and drama). This seems to be related to data 
sparsity with these two genres having fewer words than the rest as shown in Table \ref{tab:lda} and 
thus the RNNLM fine-tuning does not work very well. The experiments in Table \ref{tab:lm} were 
carried out after the challenge and thus were not a part of the final submission.

\section{System description}
\label{sec:system}

\begin{table*} [!t]
\caption{\label{tab:results} {\it Overall and individual performance results on the full 
development data set, with the Univ. of Sheffield submission for Task 1 of the MGB challenge.}}
\centerline{
\begin{tabular}{|c|c|c|c|c|c|c|c|c|c|}
\hline
System & Adv. & Child. & Comed. & Compet. & Docum. & Dram. & Even. & News & Global \\
\hline
ASR P1               & 23.1\% & 36.5\% & 45.4\% & 25.1\% & 30.0\% & 40.8\% & 36.4\% & 14.1\% & 31.2\% \\
\hline
ASR P2-1            & 22.8\% & 31.0\% & 42.9\% & 24.1\% & 28.4\% & 38.6\% & 33.6\% & 14.2\% & 29.4\% \\
ASR P2-2            & 23.0\% & 31.2\% & 42.8\% & 24.2\% & 28.5\% & 39.0\% & 33.5\% & 13.8\% & 29.4\% \\
ASR P2-3            & 23.7\% & 32.0\% & 45.3\% & 25.1\% & 29.3\% & 40.5\% & 34.3\% & 15.0\% & 30.5\% \\
\hline
System combination   & 21.6\% & 27.7\% & 40.9\% & 22.7\% & 26.6\% & 37.1\% & 31.3\% & 13.2\% & 27.5\% \\
\hline
\end{tabular}}
\end{table*}

\begin{figure}[!t]
\centering
\epsfig{figure=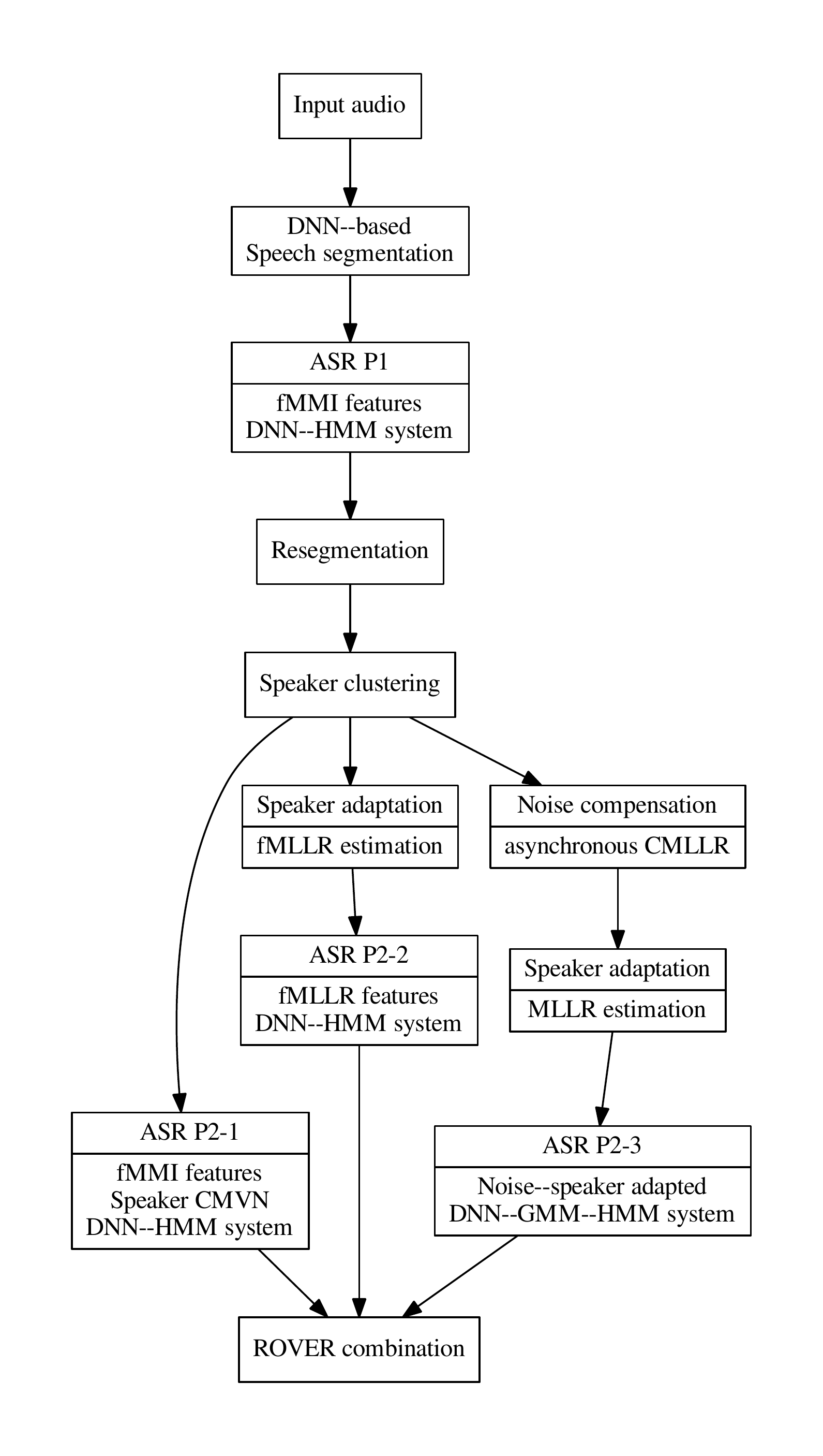,width=75mm}
\caption{{System diagram}}
\label{fig:system}
\end{figure}

The final system processing as submitted for the the MGB challenge followed the diagram pictured in 
Figure \ref{fig:system}. Each node in the diagram was implemented as a composition of separate 
modules, each performing specific computation on the speech data.

The input audio was split into speech segments using a DNN segmenter based on the $SNS2$ strategy, 
as defined in Section \ref{sec:seg}. These segments were then decoded by an initial, unadapted 
$Hybrid$ ASR system: \emph{ASR-P1}, trained on $TRN1$. The segmentation was afterwards refined using 
confidence measures in the ASR output as described in section \ref{sec:seg}. After resegmentation, 
speaker clustering based on Bayesian Information Criterion (BIC) \cite{Chen98} was performed to 
assign each speech segment to a given speaker.

From here onwards, three different decoding passes were deployed: \emph{ASR-P2-1}, \emph{ASR-P2-2} 
and \emph{ASR-P2-3}, which where based on complementary forms of dealing with speaker and noise 
variability. \emph{ASR-P2-1} was a $Hybrid$ system where the features were normalised using 
speaker-based Cepstral Mean and Variance Normalisation (CMVN) without requiring any previous 
transcript. \emph{ASR-P2-2} was also a $Hybrid$ system, but in this case speaker variability was 
compensated through the use of fMLLR input features based on the transcript from \emph{ASR-P1}. 
Finally, \emph{ASR-P2-3} was a $Bottleneck$ system where asynchronous noise transformations were used as described in 
Section \ref{sec:varia}, and speaker-based MLLR transformations were trained on top of this for 
further speaker and noise factorisation. All these three systems were trained following the sMBR 
criterion using the $TRN2$ training data definition.

The output of these three passes was finally combined via a Recognition Output Voting Error 
Reduction (ROVER) \cite{Fiscus97} procedure.
\subsection{System implementation}

The implementation of the system is based on the Resource Optimisation Toolkit (ROTK), which is 
developed by the team at the University of Sheffield and was presented initially in \cite{Hain2012}. 
ROTK allows the formulation of functional modules that can be executed in asynchronous fashion using 
computing grid infrastructure. Systems are defined as a set of modules linked together by directed 
links transferring data of specific types. This is informally depicted in a graph in Figure 
\ref{fig:system}; the actual modules used are more specific. The system uses metadata to organise 
how data is processed in an efficient parallelised way through the graph.  Each module can split its 
own tasks into several subtasks based on data, which then can be processed in parallel. The overall 
dependency structure of these sub-tasks is then automatically inferred. Each module submits jobs on 
a grid system using the Sun Grid Engine (SGE). The ROTK system allows for simple repeatability of 
the experiments as the same graph can be executed on multiple datasets such as development and 
evaluation sets.

\section{Results}
\label{sec:results}

The results of all intermediate passes and the final output are presented in Table 
\ref{tab:results}. In this Table, the gains obtained by the 3 adapted systems 
in relation to the baseline can be seen, as well as the final gain obtained by the 
combination of the three outputs. Since the results that lead to the development of the proposed
system have already been presented and discussed all through the paper, this Section only reviews
the final results achieved by the full system on the development set.

Evaluating the results per genre, the results  vary significantly from \emph{News} shows, with a 
13.2\% WER, to \emph{Comedy} shows, with a 40.9\% WER. This highlights the considerable impact
of the acoustic variability present in broadcast shows. In terms of gain, \emph{Children's} shows 
achieved the largest improvement from the initial unadapted system, 36.5\%, to the final output, 
27.7\%. This shows how the different techniques proposed for compensating variability worked in
complementary ways in one of the most challenging conditions, i.e., where children and adults may appear 
in the same show and large amounts of music and other backgrounds happen.

\section{Conclusion}
\label{sec:conclusion}

In this paper we presented the  complete system structure, model training and implementation of 
the University of Sheffield system for speech--to--text transcription of broadcast media.
The system was designed for participation in Task 1 of  the MGB challenge. The final result, 
27.5\% WER, reflects the complexity of the task, especially in the most challenging genres such as 
comedy or drama shows. It is important to note that these results are obtained without the 
availability of high quality training data, which is normally available for other related 
evaluation campaigns. The proposed system has made use of the complementarity of DNN-HMM and 
DNN-GMM-HMM systems using different adaptation strategies.

Several techniques have been proposed and evaluated. In terms of data selection techniques for acoustic model training, results have shown that
adding more data of more quality can provide improvements in both $Hybrid$ and $Bottleneck$ models. The refinement of automatic speech segmentation using
the output of an ASR stage is a significant contribution of this system, with the results showing how this can be used to find speech segments that minimise error
rates without necessarily minimising segmentation error rates. The two techniques proposed for domain and noise adaptation of acoustic models have shown
how complementary techniques can be used successfully. In this work, domain--based input features have been shown to reduce domain variability in $Hybrid$ systems;
while asynchronous adaptation with CMLLR transformations performs a similar effect in $Bottleneck$ systems. Finally, language model adaptation to multi--genre shows have
been shown to produce slight improvements. In this case, the use of genre--dependent 4--grams does not achieve the gains obtained using genre information in
RNNLMs, indicating that more work should be focused on adaptation of RNNs for language modelling.

\section{Acknowledgements and Data}
We would like to thank others in the MINI research group at Sheffield that have helped  
to develop this system, with their advice and discussions. We would also like to thank our partners 
in the NST programme, at the Universities of Cambridge and Edinburgh, for the many discussions 
which helped us greatly in the development of systems.  

The audio and subtitle data used for these experiments were distributed as part of the MGB Challenge 
(www.mgb-challenge.org) through a licence with the BBC. System output and results for the presented 
system are also available as part of the challenge results to participants.

\bibliography{biblio}
\bibliographystyle{IEEEbib}

\end{document}